\documentclass[letterpaper]{article} 
\usepackage{aaai25}  
\usepackage{times}  
\usepackage{helvet}  
\usepackage{courier}  
\usepackage[hyphens]{url}  
\usepackage{graphicx} 
\usepackage{hyperref}
\usepackage{natbib}  
\usepackage{caption} 
\let\titleold\title
\renewcommand{\title}[1]{\titleold{#1}\newcommand{\thetitle}{#1}}
\def\maketitlesupplementary
   {
   \newpage
   \begingroup 
       \centering
       \Large
       \textbf{\thetitle}\\\vspace{0.5em}
       Supplementary Material\\\vspace{1.0em}
   \endgroup 
   }

\title{Towards Adversarially Robust Dataset Distillation by Curvature Regularization}
\author{
    Eric Xue\textsuperscript{\rm 1}, 
    Yijiang Li\textsuperscript{\rm 2},
    Haoyang Liu\textsuperscript{\rm 3},
    Peiran Wang\textsuperscript{\rm 3},
    Yifan Shen\textsuperscript{\rm 4},
    Haohan Wang\textsuperscript{\rm 3}
}
\affiliations{
    \textsuperscript{\rm 1}Department of Computer Science, University of Toronto\\
    \textsuperscript{\rm 2}Electrical and Computer Engineering, University of California San Diego\\
    \textsuperscript{\rm 3}School of Information Sciences, University of Illinois Urbana-Champaign\\
    \textsuperscript{\rm 4}Siebel School of Computing and Data Science,  University of Illinois Urbana-Champaign\\

    e.xue@mail.utoronto.ca,
    yijiangli@ucsd.edu,
    whilebug@gmail.com,
    \{hl57, yifan26, haohanw\}@illinois.edu,
    
}

\usepackage{multirow}
\usepackage{amsfonts}
\usepackage{comment}
\usepackage{lipsum}
\usepackage[boxed]{algorithm2e}
\usepackage{amsmath}
\usepackage{booktabs}
\usepackage{amsthm}
\usepackage{placeins}
\usepackage{xspace}
\usepackage{subcaption}
\usepackage{enumitem}
\newtheorem{proposition}{Proposition}
\newcommand{\sre}{SRe\textsuperscript{2}L}

\newcommand{\T}{\mathcal{T}}
\newcommand{\mS}{\mathcal{S}}

\newcommand{\bx}{\textbf{x}}

\newcommand{\by}{\textbf{y}}

\newcommand{\bs}{\textbf{s}}

\newcommand{\ELL}{\mathcal{L}}

\SetKwData{KwInput}{Input:}
\renewcommand{\KwData}{\textbf{Input: }}
\renewcommand{\KwResult}{\textbf{Output: }}
\SetKwFor{ForEach}{for each}{do}{end for}
\DeclareMathOperator*{\argmin}{arg\,min}
\newcommand{\setlabel}[1]{\edef\@currentlabel{#1}\label}

\begin{document}

\maketitle

\begin{abstract}
Dataset distillation (DD) allows datasets to be distilled to fractions of their original size while preserving the rich distributional information, so that models trained on the distilled datasets can achieve a comparable accuracy while saving significant computational loads. Recent research in this area has been focusing on improving the accuracy of models trained on distilled datasets. In this paper, we aim to explore a new perspective of DD. We study how to embed adversarial robustness in distilled datasets, so that models trained on these datasets maintain the high accuracy and meanwhile acquire better adversarial robustness. We propose a new method that achieves this goal by incorporating curvature regularization into the distillation process with much less computational overhead than standard adversarial training. Extensive empirical experiments suggest that our method not only outperforms standard adversarial training on both accuracy and robustness with less computation overhead but is also capable of generating robust distilled datasets that can withstand various adversarial attacks. Our implementation is available at: \href{https://github.com/yumozi/GUARD}{https://github.com/yumozi/GUARD}.
\end{abstract}

%

\section{Introduction}
\label{sec: intro}
In the era of big data, the computational demands for training deep learning models are continuously growing due to the increasing volume of data. This presents substantial challenges, particularly for entities with limited computational resources. To mitigate such issues, concepts like dataset distillation~\cite{wang2018distillation} and dataset condensation~\cite{zhao2021dc, zhao2021dsa, zhao2023dm} have emerged, offering a means to reduce the size of the data while maintaining its utility. A successful implementation of dataset distillation can bring many benefits, such as enabling more cost-effective research on large datasets and models.

Dataset distillation (DD) refers to the task of synthesizing a smaller dataset such that models trained on this smaller set yield high performance when tested against the original, larger dataset. Dataset distillation algorithms take a large dataset as input and generate a compact, synthetic dataset. The efficacy of the distilled dataset is assessed by evaluating models trained on it against the original dataset.

Conventionally, distilled datasets are evaluated based on their standard test accuracy. Therefore, recent research has expanded rapidly in the direction of improving the test accuracy following the evaluation procedure \cite{sachdeva2023survey}. Additionally, many studies focus on improving the efficiency of the distillation process \cite{sachdeva2023survey}.

Less attention, however, has been given to an equally important aspect of this area of research: the adversarial robustness of models trained on distilled datasets. Adversarial robustness is a key indicator of a model's resilience against malicious inputs, making it a crucial aspect of trustworthy machine learning. Given the potential of dataset distillation to safeguard the privacy of the original dataset \cite{geng2023survey, chen2023survey}, exploring its capability to also enhance model robustness opens a promising avenue for advancing research in trustworthy machine learning \cite{liu2023trustworthy}. Thus, our work seeks to bridge this gap and focuses on the following question: \textbf{How can we embed adversarial robustness into the dataset distillation process, thereby generating datasets that lead to more robust models?}

Motivated by this question, we explore potential methods to accomplish this goal. As it turns out, it is not as simple as adding adversarial training to the distillation process. To find a more consistent method, we study the theoretical connection between adversarial robustness and dataset distillation. Our theory suggests that we can directly improve the robustness of the distilled dataset by minimizing the curvature of the loss function with respect to the real data. Based on our findings, we propose a novel method, GUARD (\textbf{G}eometric Reg\textbf{u}larization for \textbf{A}dversarially \textbf{R}obust \textbf{D}ataset), which incorporates curvature regularization into the distillation process. We then evaluate GUARD against existing distillation methods on ImageNette, Tiny ImageNet, and ImageNet datasets. In summary, the contributions of this paper are as follows
\begin{itemize}
    \item Empirical and theoretical exploration of adversarial robustness in distilled datasets
    \item A theory-motivated method, GUARD, that offers robust dataset distillation with minimal computational overhead
    \item Detailed evaluation of GUARD to demonstrate its effectiveness across multiple aspects
\end{itemize}

\section{Related Works}
\setlabel{Related Works}{sec: related_works}
\subsection{Dataset Distillation}
Aiming to address the issue of the increasing amount of data required to train deep learning models, the goal of dataset distillation is to efficiently train neural networks using a small set of synthesized training examples from a larger dataset. Dataset distillation (DD)~\cite{wang2018distillation} was one of the first such methods developed, and it showed that training on a few synthetic images can achieve similar performance on MNIST and CIFAR10 as training on the original dataset. Later, \citet{cazenavette2022mtt,zhao2021dsa,zhao2021dc, pmlr-v162-lee22b} explored different methods of distillation, including gradient and trajectory matching  w.r.t. the real and synthetic data, with stronger supervision for the training process. Instead of matching the weights of the neural network, another thread of works~\cite{9879629, zhao2023dm, zhang2024m3d, liu2023dataset} focuses on matching feature distributions of the real and synthetic data in the embedding space to better align features or preserve real-feature distribution. Considering the lack of efficiency of the bi-level optimization in previous methods, \citet{nguyen2021dataset, zhou2022frepo} aim to address the significant amount of meta gradient computation challenges. \citet{nguyen2020dataset} proposed a kernel-inducing points meta-learning algorithm and they further leverage the connection between the infinitely wide ConvNet and kernel ridge regression for better performance. Furthermore, \citet{sucholutsky2021soft} addresses the simultaneous distillation of images and their corresponding soft labels. Later, some works focused on further improving efficiency of the process, such as \citet{yin2023squeeze} that introduced \sre, which optimizes the distillation process by dividing it into three distinct steps for greater efficiency, and \citet{xu2024distill}, which proposed an approach to enhance both the efficiency and performance by first pruning the original dataset. Finally, \citet{li2024alignment} further advanced the process by dynamically pruning the original dataset based on the desired compression ratio and extracting information from deeper layers of the network.

Dataset distillation approaches can be broadly classified into four families based on their underlying principles: meta-model matching, gradient matching, distribution matching, and trajectory matching~\cite{sachdeva2023survey}. Regardless of the particular approach, most of the existing methods rely on optimizing the distilled dataset w.r.t. a network trained with real data, such methods include DD~\cite{wang2018distillation}, DC~\cite{zhao2021dc}, DSA~\cite{zhao2021dsa}, MTT~\cite{cazenavette2022mtt}, DCC~\cite{pmlr-v162-lee22b}, \sre~\cite{yin2023squeeze}, ATT~\cite{liu2024att} and many more. 

In a related direction, some works also address the robustness of dataset distillation, but specifically focusing on out-of-distribution (OOD) robustness. For instance, \citet{vahidian2024risk} employs risk minimization techniques to ensure robustness, while TrustDD~\cite{ma2024trustworthy} incorporates outliers during the distillation process to facilitate OOD detection.

\subsection{Adversarial Attacks}
Adversarial attacks are a significant concern in the field of machine learning, as they can cause models to make incorrect predictions even when presented with seemingly similar input.  \citet{kurakin2017adversarial} demonstrates the real-world implications of these attacks. Many different types of adversarial attacks have been proposed in the literature~\cite{goodfellow2015adversarial, madry2018adversarial}. In particular, Projected Gradient Descent (PGD) is a widely used adversarial attack that has been shown to be highly effective against a variety of machine learning models~\cite{madry2018adversarial}. The limitations of defensive distillation, a technique initially proposed for increasing the robustness of machine learning models, were explored by~\citet{papernot2016limitations}. \citet{moosavi2016deepfool} introduced DeepFool, an efficient method to compute adversarial perturbations. Other notable works include the study of the transferability of adversarial attacks by \citet{papernot2016transferability}, the simple and effective black-box attack by \citet{narodytska2016simple}, and the zeroth-order optimization-based attack by \citet{chen2017zoo}. More recently, \citet{athalye2018obfuscated} investigated the robustness of obfuscated gradients, and \citet{wong2019wasserstein} introduced the Wasserstein smoothing as a novel defense against adversarial attacks. \citet{croce2020adversarial} introduced AutoAttack, which is a suite of adversarial attacks consisting of four diverse and parameter-free attacks that are designed to provide a comprehensive evaluation of a model's robustness to adversarial attacks.

\subsection{Adversarial Defense}
Numerous defenses against adversarial attacks have been proposed. Among these, adversarial training stands out as a widely adopted defense mechanism that entails training machine learning models on both clean and adversarial examples~\cite{goodfellow2015adversarial}. Several derivatives of the adversarial training approach have been proposed, such as ensemble adversarial training~\cite{tramer2018adversarial}, and randomized smoothing~\cite{cohen2019adversarial}.
However, while adversarial training can be effective, it bears the drawback of being computationally expensive and time-consuming.

Some defense mechanisms adopt a geometrical approach to robustness. One such defense mechanism is CURE ~\cite{moosavi2019robustness}, a method that seeks to improve model robustness by reducing the curvature of the loss landscape during training. 
Similarly, \citet{miyato2015distributional} improved the smoothness of the output distribution, \citet{cisse2017parseval} enforced Lipschitz constants, \citet{ross2018improving} employed input gradient regularization, to improve the models' adversarial robustness.  

Several other types of defense techniques have also been proposed, such as corrupting with additional noise and pre-processing with denoising autoencoders by \citet{gu2014deep}, the defensive distillation approach by \citet{papernot2016distillation}, the Houdini adversarial examples by \citet{cisse2017houdini}, and the approximate null space augmentation by \citet{liu2025approximate}.

\section{Preliminary}
\label{sec: prelim}
\subsection{Dataset Distillation}
\label{sec: background_dc}
Before we delve into the theory of robustness in dataset distillation methods, we will formally introduce the formulation of dataset distillation in this section. 

\paragraph{Notations} Let $\mathcal{T}$ represent the real dataset, drawn from the distribution $\mathcal{D_T}$. The dataset $\mathcal{T}$ comprises $n$ image-label pairs, defined as $\mathcal{T} = \{(\mathbf{x}_i, y_i)\}_{i=1}^n$. Similarly, let $\mathcal{S}$ denote the distilled dataset, drawn from the distribution $\mathcal{D_S}$, and consisting of $m$ image-label pairs, defined as $\mathcal{S} = \{(\tilde{\mathbf{x}}_j, \tilde{y}_j)\}_{j=1}^m$, where  $m \ll n$.  Conventionally, instead of directly expressing the size of the distilled dataset as $|\mathcal{S}|$, it is more common to describe it in terms of ``images per class'' (IPC). Let $\ell(\mathbf{x}, y; \boldsymbol{\theta})$ denote the loss function of a model parameterized by $\boldsymbol{\theta}$ on a sample $(\mathbf{x}, y)$, and $\mathcal{L}(\mathcal{T}; \boldsymbol{\theta})$ denotes the empirical loss on $\mathcal{T}$, $\mathcal{L}(\mathcal{T}; \boldsymbol{\theta}) = \frac{1}{n} \sum_{i=1}^{n} \ell(\mathbf{x}_i, y_i; \boldsymbol{\theta})$.

Given the real training set $\mathcal{T}$, dataset distillation aims to find the optimal synthetic dataset $\mathcal{S}^*$ by solving the following bi-level optimization problem:

\begin{equation}
\begin{gathered}
\label{eq: formulation}
    \mathcal{S}^* = \argmin_{\mathcal{S}}  \mathop{\mathbb{E}}_{{(\mathbf{x}, y)} \sim \mathcal{D_T}} \ell\left(\mathbf{x}, y; \boldsymbol{\theta}(\mathcal{S})\right) \\
    \textrm{subject to}~~ \boldsymbol{\theta}(\mathcal{S}) = \argmin_{\boldsymbol{\theta}} \mathcal{L}(\mathcal{S}; \boldsymbol{\theta}).
\end{gathered}
\end{equation}

Directly solving this problem requires searching for the optimal parameters in the inner problem and unrolling the gradient descent steps in the computation graph to find the hypergradient with respect to $\mathcal{S}$, which is computationally expensive. One common alternative approach is to align a model trained on the distilled set with one trained on the real dataset. Conceptually, it can be summarized in the below equation:
\begin{equation}
\begin{gathered}
\min_\mathcal{S}D
({\boldsymbol{\theta}(\mathcal{S})}, {\boldsymbol{\theta}(\mathcal{T})}) \\
\textrm{subject to}\quad \boldsymbol{\theta}(\mathcal{S}) = \argmin_{\boldsymbol{\theta}} \mathcal{L}(\mathcal{S}; \boldsymbol{\theta})
\\ \textrm{and} \quad\boldsymbol{\theta}(\mathcal{T}) =  \argmin_{\boldsymbol{\theta}} \mathcal{L}(\mathcal{T}; \boldsymbol{\theta}),
\end{gathered}
\end{equation}
where $D$ is a manually chosen distance function. Recent works have proliferated along this direction, with methods such as gradient matching~\cite{zhao2021dc} and trajectory matching~\cite{cazenavette2022mtt}, each focusing on aligning different aspects of the model's optimization dynamics. 
Some works have also tried to align the distribution of the distilled data with that of the real data \cite{zhao2023dm, zhang2024m3d, liu2023dataset}, or recover a distilled version of the training data from a trained model \cite{yin2023squeeze, buzaglo2023data}. These methods do not rely on the computation of second-order gradients, leading to improved efficiency and performance on large-scale datasets.

Despite the wide spectrum of methods for dataset distillation, they were primarily designed for improving the standard test accuracy, and significantly less attention has been paid to the adversarial robustness. In the following, we conduct a preliminary study to show that adversarial robustness cannot be easily incorporated into the distilled data by the common approach of adversarial training, necessitating more refined analysis.

\subsection{The Limitation of Adversarial Training in Dataset Distillation} 
\label{sec:adv_training}
\begin{table}[ht]
\centering
\fontsize{9pt}{11pt}\selectfont
\footnotesize
\caption{Accuracy of ResNet18 on ImageNette trained on distilled datasets from GUARD, \sre~, and \sre~with adversarial training}
\setlength{\tabcolsep}{2mm}{
\begin{tabular}{ccccc}
\toprule
IPC                 & Attack       & GUARD & \sre & \sre~+Adv \\ \midrule
\multirow{6}{*}{1}  & None (Clean) & \textbf{37.49} & 27.97 & 11.61                \\
                    & PGD100       & \textbf{16.22} & 12.05 & 10.03                \\
                    & Square       & \textbf{26.74} & 18.62 & 11.18                \\
                    & AutoAttack   & \textbf{15.81} & 12.12 & 10.03                \\
                    & CW           & \textbf{29.14} & 20.38 & 10.31                \\
                    & MIM          & \textbf{16.32} & 12.05 & 10.03                \\ \midrule
\multirow{6}{*}{10} & None (Clean) & \textbf{57.93} & 42.42 & 12.81                \\
                    & PGD100       & \textbf{23.87} & 4.76 & 9.93                 \\
                    & Square       & \textbf{44.07} & 22.77 & 11.46                \\
                    & AutoAttack   & \textbf{19.69} & 4.99 & 9.96                 \\
                    & CW           & \textbf{58.67} & 22.11 & 10.90                \\
                    & MIM          & \textbf{21.80} & 4.76 & 9.96                 \\ \bottomrule
\end{tabular}
}
\label{table:adv_teaser}
\end{table}

In the supervised learning setting, one of the most commonly used methods to enhance model robustness is adversarial training, which involves training the model on adversarial examples that are algorithmically searched for or crafted \cite{goodfellow2015adversarial}. This can be formulated as  
\begin{align}
\min_{\boldsymbol{\theta}} \mathop{\mathbb{E}}_{(\mathbf{x}, y) \sim \mathcal{D}} \left(\max_{\|\mathbf{v}\| \le \rho} \ell(\mathbf{x} + \mathbf{v}, y; \boldsymbol{\theta})\right),
\end{align}
where $\mathbf{v}$ is some perturbation within the $\ell_p$ ball with radius $\rho$, and $\mathcal{D}$ is the data distribution.

Analogously, in the dataset distillation setting, one intuitive way to distill robust datasets would be to synthesize a distilled dataset using a robust model trained with adversarial training. As mentioned in the related works section, many dataset distillation methods utilize a model trained on the original dataset as a comparison target, therefore this technique can be easily integrated to those methods. 

While embedding adversarial training directly within the dataset distillation process may seem like an intuitive and straightforward approach, our comprehensive analysis reveals its limitations across various distillation methods. As an example, we show the evaluation of one such implementation based on \sre~\cite{yin2023squeeze} in Table \ref{table:adv_teaser}. The results indicate a significant decline in clean accuracy for models trained on datasets distilled using this technique, in contrast to those synthesized by the original method. Moreover, the improvements in robustness achieved are very inconsistent. 
In our experiment, we only employed a weak PGD attack with $\epsilon = 1/255$ to generate adversarial examples for adversarial training, leading to the conclusion that even minimal adversarial training can detrimentally impact model performance when integrated into the dataset distillation process.

Such outcomes are not entirely unexpected. Previous studies, such as those by~\citet{zhang2020attacks}, have indicated that adversarial training can significantly alter the semantics of images through perturbations, even when adhering to set norm constraints. This can lead to the cross-over mixture problem, severely degrading the clean accuracy. We hypothesize that these adverse effects might be magnified during the distillation process, where the distilled dataset's constrained size results in a distribution that is vastly different from that of the original dataset.

\section{Methods}
\setlabel{Methods}{sec: methods}
\subsection{Formulation of the Robust Distillation Problem} 
Extending the distillation problem to the adversarial robustness setting, robust dataset distillation can be formulated as a tri-level optimization problem as below:
\begin{equation}
\begin{gathered}
\label{eq: formulation_robust}
    \mathcal{S}^* = \argmin_{\mathcal{S}} \mathop{\mathbb{E}}_{{(\mathbf{x}, y)} \sim \mathcal{D_T}} \left(\max_{\|\mathbf{v}\| \le \rho} \ell\left(\mathbf{x} + \mathbf{v}, y; \boldsymbol{\theta}(\mathcal{S})\right)\right) \\
    \textrm{subject to}~~ \boldsymbol{\theta}(\mathcal{S}) = \argmin_{\boldsymbol{\theta}} \mathcal{L}(\mathcal{S}; \boldsymbol{\theta}).
\end{gathered}
\end{equation}
If we choose to directly optimize for the robust dataset distillation objective, the tri-level optimization problem will result in a hugely inefficient process. Instead, we will uncover a theoretical relationship between dataset distillation and adversarial robustness to come up with a more efficient method that avoids the tri-level optimization process.

\subsection{Theoretical Bound of Robustness}
\label{theory}
Our aim is to create a method that allows us to efficiently and reliably introduce robustness into distilled datasets, thus we will start by exploring the theoretical connections between dataset distillation and adversarial robustness. Conveniently, previous works~\cite{Jetley_Lord_Torr_2018, Fawzi_2018} have studied the adversarial robustness of neural networks via the geometry of the loss landscape. Inspired by~\citet{moosavi2016deepfool}, here we find connections between standard training procedures and dataset distillation to provide a theoretical bound for the adversarial loss of models trained with distilled datasets.

Let $\ell(\mathbf{x}, y; \theta)$ denote the loss function of the neural network, or $\ell(\mathbf{x})$ for simplicity, and $\mathbf{\mathbf{v}}$ denote a perturbation vector. By Taylor's Theorem,
\begin{align}
\ell(\mathbf{x+v}) = \ell(\mathbf{x}) + {\nabla\ell(\mathbf{x})}^\top \mathbf{v} + \frac{1}{2} \mathbf{v^\top Hv} + o({\| \mathbf{v} \|}^2).
\end{align}
We are interested in the property of $\ell(\cdot)$ in the locality of $\mathbf{x}$, so we focus on the quadratic approximation $\tilde{\ell}(\mathbf{x+v}) = \ell(\mathbf{x}) + {\nabla\ell(\mathbf{x})}^\top \mathbf{v} + \frac{1}{2} \mathbf{v^\top Hv}$. We define the adversarial loss on real data as $\tilde{\ell}_\rho^{adv}(\mathbf{x}) = \max_{\|\mathbf{v}\| \le \rho} \tilde{\ell}(\mathbf{x+v})$. We can expand this and take the expectation over the distribution with class label $c$, denoted as $D_c$, to get the following:
\begin{align}
\label{eq:exp_taylor_main}
\begin{gathered}
\mathop{\mathbb{E}}_{\mathbf{x} \sim D_c} \tilde{\ell}_\rho^{adv}(\mathbf{x}) \le 
\mathop{\mathbb{E}}_{\mathbf{x} \sim D_c}\ell(\mathbf{x}) + \rho\mathop{\mathbb{E}}_{\mathbf{x} \sim D_c} \| {\nabla \ell(\mathbf{x})}\| + \\ \frac{1}{2} \rho^2 \mathop{\mathbb{E}}_{\mathbf{x} \sim D_c} \lambda_1(\mathbf{x}), 
\end{gathered}
\end{align}
where $\lambda_1$ is the largest eigenvalue of the Hessian matrix $\mathbf{H(\ell(x))}$.  Then, we have the proposition:
\begin{proposition}
Let $\mathbf{x}^\prime$ be a distilled datum with the label $c$ and satisfies $\|h(\mathbf{x}^\prime) - \mathbb{E}_{\mathbf{x} \sim D_c}[h(\mathbf{x})]\| \le \sigma$, where $h(\cdot)$ is a feature extractor. Assume $\ell(\cdot)$ is convex in $\mathbf{x}$ and $\tilde{\ell}_\rho^{adv}(\cdot)$ is $L$-Lipschitz in the feature space, then the below inequality holds:
\begin{align}
\label{eq:bound_main}
\begin{gathered}
    \tilde{\ell}_\rho^{adv}(\mathbf{x}^\prime) \le \mathop{\mathbb{E}}_{\mathbf{x} \sim D_c} \ell(\mathbf{x}) + \rho \mathop{\mathbb{E}}_{\mathbf{x} \sim D_c} \|\nabla \ell(\mathbf{x})\| + \\ \frac{1}{2} \rho^2 \mathop{\mathbb{E}}_{\mathbf{x} \sim D_c} \lambda_1(\mathbf{x}) + L\sigma.
\end{gathered}
\end{align}
\end{proposition}
Given the assumption of convexity in the loss function, we can further observe that in a convex landscape the gradient magnitude tends to be lower near the optimal points. Therefore, in the context of a convex loss function and a well-distilled dataset, the gradient term $\rho \mathop{\mathbb{E}}_{\mathbf{x} \sim D_c} \|\nabla \ell(\mathbf{x})\|$ contribute insignificantly to RHS of Eq. \ref{eq:bound_main}. This insignificance is amplified by the presence of the curvature term, $\frac{1}{2} \rho^2 \mathop{\mathbb{E}}_{\mathbf{x} \sim D_c} \lambda_1(\mathbf{x})$, which provides a sufficient descriptor of the loss landscape under our assumptions. Hence, it is reasonable to simplify the expression by omitting the gradient term, resulting in a focus on the curvature term, which is more representative of the convexity assumption and the characteristics of a well-distilled dataset. The revised expression would then be:
\begin{align}
\label{eq:bound_simple}
    \tilde{\ell}_\rho^{adv}(\mathbf{x}^\prime) \le \mathop{\mathbb{E}}_{\mathbf{x} \sim D_c} \ell(\mathbf{x}) + \frac{1}{2} \rho^2 \mathop{\mathbb{E}}_{\mathbf{x} \sim D_c} \lambda_1(\mathbf{x}) + L\sigma.
\end{align}

Dataset distillation methods usually already optimizes for $\ell(\mathbf{x})$, and we can also assume that the $\sigma$ for a well-distilled dataset is small. Hence, we can conclude that the upper bound of adversarial loss of distilled datasets is largely affected by the curvature of the loss function in the locality of real data samples.

In the appendix, we give a more thorough proof of the proposition and discuss the validity of some of the assumptions made. In the Experiments section, we also show results from an ablation study to demonstrate the empirical effects of some of these assumptions.
\subsection{Geometric Regularization for Adversarially Robust Dataset}

Based on our theoretical discussion, we propose a method, GUARD (\textbf{G}eometric Reg\textbf{u}larization for \textbf{A}dversarially \textbf{R}obust \textbf{D}ataset). Since the theorem suggests that the upper bound of the adversarial loss is mainly determined by the curvature of the loss function, we modify the distillation process so that the trained model has a loss function with a low curvature with respect to real data.

Reducing $\lambda_1$ in Eq. \ref{eq:bound_simple} requires computing the Hessian matrix to get the largest eigenvalue $\lambda_1$, which is quite computationally expensive. Here we find an efficient approximation of it. Let $\mathbf{v_1}$ be the unit eigenvector corresponding to $\lambda_1$, then the Hessian-vector product is
\begin{align}
    \mathbf{Hv_1} = \lambda_1 \mathbf{v_1} = \lim_{h \to 0} \frac{\nabla \ell(\mathbf{x}+h\mathbf{v}_1) - \nabla\ell(\mathbf{x})}{h}. 
\end{align}
We take the differential approximation of the Hessian-vector product, because we are interested in the curvature in a local area of $x$ rather than its asymptotic property. Therefore, for a small $h$, 
\begin{align}
    \lambda_1 = \|\lambda_1\mathbf{v_1}\| \approx \| \frac{\nabla \ell(\mathbf{x}+h\mathbf{v}_1) - \nabla\ell(\mathbf{x})}{h} \|.
\end{align}
Previous works~\cite{Fawzi_2018, Jetley_Lord_Torr_2018, moosavi2019robustness} have empirically shown that the direction of the gradient has a large cosine similarity with the direction of $\mathbf{v_1}$ in the input space of neural networks. Instead of calculating $\mathbf{v_1}$ directly, it is more efficient to take the gradient direction as a surrogate of $\mathbf{v_1}$ to perturb the input $\mathbf{x}$. So we replace the $\mathbf{v}_1$ above with the normalized gradient $\mathbf{z} = \frac{\nabla\ell(\mathbf{x}))}{\|\nabla\ell(\mathbf{x}))\|}$, and define the regularized loss $\ell_R$ to encourage linearity in the input space:
\begin{align}
\label{rloss_main}
    \ell_R(\mathbf{x}) = \ell(\mathbf{x}) + \lambda \| \nabla \ell(\mathbf{x}+h\mathbf{z}) - \nabla\ell(\mathbf{x}) \|^2,
\end{align}
where $\ell$ is the original loss function, $h$ is the discretization step, and the denominator $h$ is merged with the regularization coefficient $\lambda$.

\subsection{Engineering Specification}
\label{sec:engineering}
In order to evaluate the effectiveness of our method, we implemented GUARD using the \sre~method as a baseline. We incorporated our regularizer into the squeeze step of \sre~by substituting the standard training loss with the modified loss outlined in Eq. \ref{rloss_main}. In the case of \sre, this helps to synthesize a robust distilled dataset by allowing images to be recovered from a robust model in the subsequent recover step.

\section{Experiments}
\label{sec: exp}
\subsection{Experiment Settings}
For a systematic evaluation of our method, we investigate the top-$1$ classification accuracy of models trained on data distilled from three commonly-used datasets in this area: ImageNette \cite{imagenette}, Tiny ImageNet \cite{le2015tiny}, and ImageNet-1K \cite{deng2009imagenet}. ImageNette is a subset of ImageNet-1K containing 10 easy-to-classify classes. Tiny ImageNet is a scaled-down subset of ImageNet-1K, containing 200 classes and 100,000 downsized 64x64 images. We trained networks using the distilled datasets and subsequently evaluated the network's performance on the validation split of the original datasets (because none of these datasets have a test split with publicly available labels). For consistency in our experiments across all datasets, we used the standard ResNet18 architecture~\cite{he2016resnet} to synthesize the distilled datasets and evaluate their performance.

During the squeeze step of the distillation process, we trained the model on the original dataset over 50 epochs using a learning rate of 0.025. Based on preliminary experiments, we determined that the settings $h=3$ and $\lambda=100$ provide an optimal configuration for our regularizer. In the recover step, we performed 2000 iterations to synthesize the images and run 300 epochs to generate the soft labels to obtain the full distilled dataset. In the evaluation phase, we trained a ResNet18 model on the distilled dataset for 300 epochs, before assessing it on the validation split of the original dataset.

\subsection{Comparison with Other Methods}
As of now, there is only a small number of dataset distillation methods that can achieve good performance on ImageNet-level datasets, therefore our choices for comparison is small. Here, we first compare our method to the original \sre~\cite{yin2023squeeze} to observe the direct effect of our regularizer on the adversarial robustness of the trained model. We also compare with MTT~\cite{cazenavette2022mtt} and TESLA~\cite{cui2023tesla} on the same datasets to gain a further understanding on the differences in robustness between our method and other dataset distillation methods. We utilized the exact ConvNet architecture described in the papers of MTT and TESLA for their distillation and evaluation, as their performance on ResNet seems to be significantly lower.

We evaluate all the methods on three distillation scales: 10 IPC, 50 IPC, and 100 IPC. We also employed a range of attacks to evaluate the robustness of the model, including PGD100~\cite{madry2017towards}, Square~\cite{andriushchenko2020square}, AutoAttack~\cite{croce2020adversarial}, CW~\cite{carlini2017towards}, and MIM~\cite{dong2017mim}. This assortment includes both white-box and black-box attacks, providing a comprehensive evaluation of GUARD. For all adversarial attacks, with the exception of CW attack, we use the setting $\epsilon = 1/255$. For CW specifically, we set the box constraint $c$ to $10^{-5}$. Due to computational limits, we were not able to obtain results for MTT and TESLA with the 100 IPC setting on ImageNet, as well as the 100 IPC setting on ImageNet for all methods.

\subsection{Results}
\begin{figure}[ht]
  \centering
  \includegraphics[width=0.45\textwidth]{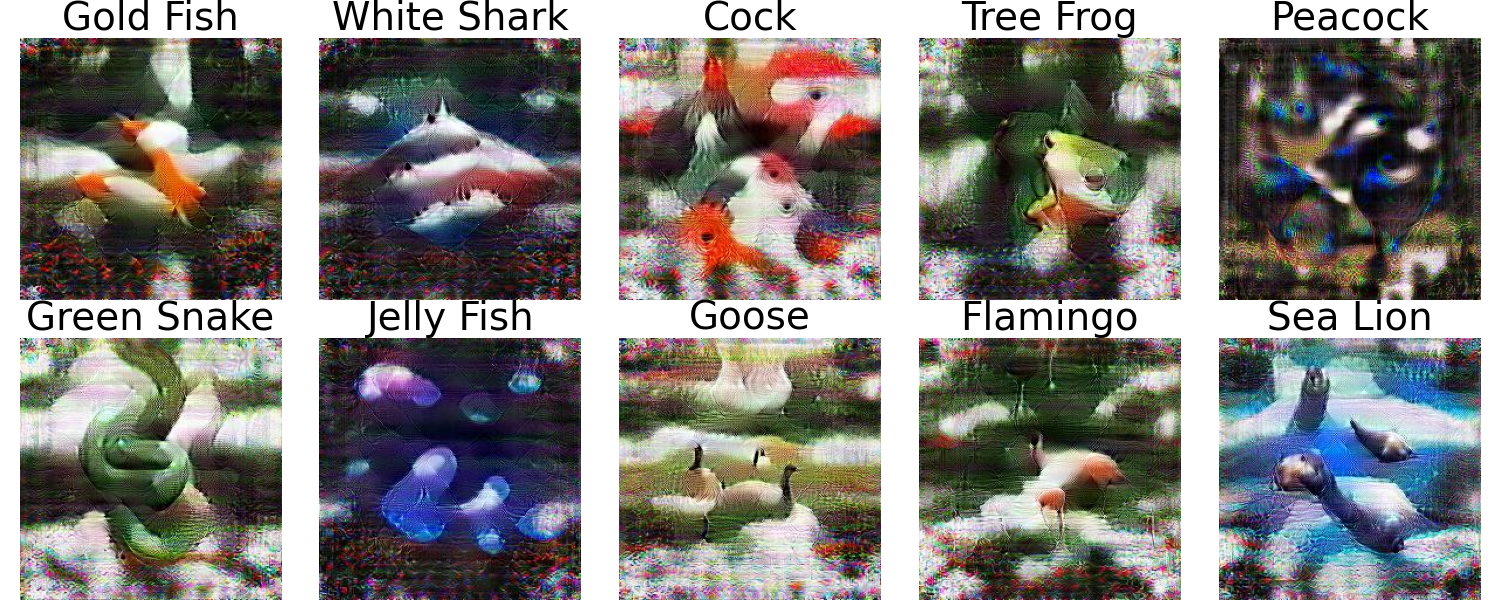}
  \caption{Visualization of distilled images generated using GUARD with 1 IPC setting from ImageNet.}
  \label{figure2}
\end{figure}

The results are detailed in Table \ref{table:main}.  It can be observed that GUARD consistently outperforms both \sre~and MTT in terms of robustness across various attacks. Interestingly, we observe an increase in clean accuracy upon incorporating GUARD across various settings. While enhancing clean accuracy was not the primary goal of GUARD, this outcome aligns with its function as a regularizer, potentially aiding in model generalization. In the context of dataset distillation, where the goal is to distill essential features of the original dataset into a smaller subset, improving the generalization is expected to have positive effects on the performance. We also provide a visualization of the distilled images generated by GUARD in Figure \ref{figure2}, utilizing a distillation scale of 1 image per class among selected ImageNet classes. It can be seen that the images exhibit characteristics that resemble a blend of multiple objects within their assigned class, highlighting the method's capacity to capture essential features.

\begin{table}[!ht]
\centering
\caption{Evaluation of different dataset distillation methods under adversarial attacks on ImageNette, TinyImageNet, and ImageNet. The best results among all methods are highlighted in bold, second best are underlined.} 
\footnotesize 
\setlength{\tabcolsep}{.5mm}{
\begin{tabular}{@{}llccccc@{}}
\toprule
\multicolumn{1}{c}{\multirow{2}{*}{Dataset}} & \multicolumn{1}{c}{\multirow{2}{*}{IPC}} & \multirow{2}{*}{Attack}          & \multicolumn{4}{c}{Methods}            \\ \cmidrule(l){4-7} 
\multicolumn{1}{c}{} & \multicolumn{1}{c}{} &                                  & GUARD          & \sre & MTT   & \multicolumn{1}{l}{TESLA} \\ \midrule
\multirow{18}{*}{ImageNette}                 & \multirow{6}{*}{10}                      & \multicolumn{1}{l}{None (Clean)} & \underline{57.93} & 42.42 & \textbf{58.43} & 36.84 \\
                     &                      & PGD100                           & 23.87          & 4.76  & \textbf{39.85} & \underline{28.10}                     \\
                     &                      & Square                           & \textbf{44.07} & 22.77 & \underline{34.79} & 24.61                     \\
                     &                      & AutoAttack                       & 19.69          & 4.99  & \textbf{33.72} & \underline{24.48}                     \\
                     &                      & CW                               & \textbf{41.47} & 22.11 & \underline{34.57} & 24.61                     \\
                     &                      & MIM                              & 21.80          & 4.76  & \textbf{39.20} & \underline{28.15}                     \\ \cmidrule(l){2-7} 
                     & \multirow{6}{*}{50}  & \multicolumn{1}{l}{None (Clean)} & \textbf{80.86} & \underline{80.15} & 59.69 & 36.21                         \\
                     &                      & PGD100                           & \textbf{41.42} & 12.30 & \underline{41.13} & 28.72                         \\
                     &                      & Square                           & \textbf{72.81} & \underline{61.50} & 36.72 & 25.34                         \\
                     &                      & AutoAttack                       & \textbf{42.47} & 12.91 & \underline{35.46} & 27.21                         \\
                     &                      & CW                               & \textbf{58.67} & \underline{53.42} & 36.54 & 29.01                         \\
                     &                      & MIM                              & \textbf{43.23} & 12.43 & \underline{41.69} & 30.12                         \\ \cmidrule(l){2-7} 
                     & \multirow{6}{*}{100} & \multicolumn{1}{l}{None (Clean)} & \underline{83.39}          & \textbf{85.83} & 64.33 & 45.04                     \\
                     &                      & PGD100                           & \textbf{57.50}          & 31.65 & \underline{44.89} & 33.98                     \\
                     &                      & Square                           & \textbf{77.68} & 19.18 & \underline{40.41} & 29.27                     \\
                     &                      & AutoAttack                       & \textbf{64.84} & 17.93 & \underline{39.46} & 28.99                     \\
                     &                      & CW                               & \textbf{69.35} & \underline{68.20} & 40.66 & 29.32                     \\
                     &                      & MIM                              & \textbf{65.07} & 18.98 & \underline{44.89} & 33.98                     \\ \midrule \midrule
\multirow{18}{*}{TinyImageNet}               & \multirow{6}{*}{10}                      & \multicolumn{1}{l}{None (Clean)} & \textbf{37.00} & \underline{33.18} & 8.14  & 14.06 \\
                     &                      & PGD100                           & \underline{6.39}           & 1.08  & 4.08  & \textbf{8.40}                      \\
                     &                      & Square                           & \textbf{19.53} & \underline{15.85} & 2.48  & 6.31                      \\
                     &                      & AutoAttack                       & \underline{4.91}           & 0.79  & 2.44  & \textbf{6.16}                      \\
                     &                      & CW                               & \textbf{8.40}  & 3.24  & 2.50  & \underline{6.26}                      \\
                     &                      & MIM                              & \underline{6.51}           & 1.10  & 4.08  & \textbf{8.40}                      \\ \cmidrule(l){2-7} 
                     & \multirow{6}{*}{50}  & \multicolumn{1}{l}{None (Clean)} & \underline{55.61}          & \textbf{56.42} & 17.84 & 28.24                         \\
                     &                      & PGD100                           & \textbf{15.63} & 0.27  & 5.62  & \underline{12.12}                        \\
                     &                      & Square                           & \textbf{36.93} & \underline{15.50} & 3.84  & 10.39                         \\
                     &                      & AutoAttack                       & \textbf{13.84} & 0.16  & 3.52  & \underline{10.01}                         \\
                     &                      & CW                               & \textbf{20.46} & \underline{12.12} & 3.66  & 10.13                         \\
                     &                      & MIM                              & \textbf{16.09} & 0.29  & 5.64  & \underline{12.12}                        \\ \cmidrule(l){2-7} 
                     & \multirow{6}{*}{100} & \multicolumn{1}{l}{None (Clean)} & \textbf{60.13} & \underline{59.30} & 29.16 & 30.48                        \\
                     &                      & PGD100                           & \underline{13.79} & 0.25  & 8.63  & \textbf{14.45}                     \\
                     &                      & Square                           & \textbf{37.06} & \underline{17.74} & 7.29  & 12.02                         \\
                     &                      & AutoAttack                       & \textbf{12.76} & 0.19  & 6.75  & \underline{11.57}                         \\
                     &                      & CW                               & \textbf{20.05} & \underline{14.02} & 6.93  & 11.57                         \\
                     &                      & MIM                              & \underline{14.35} & 0.24  & 8.63  & \textbf{14.45}                        \\ \midrule \midrule
\multirow{12}{*}{ImageNet-1K}                   & \multirow{6}{*}{10}                      & \multicolumn{1}{l}{None (Clean)} & \textbf{27.25} & 21.30 & -     & -     \\
                     &                      & PGD100                           & \textbf{5.25}  & \underline{0.55}  & -     & -                         \\
                     &                      & Square                           & \underline{17.88}          & \textbf{18.02} & -     & -                         \\
                     &                      & AutoAttack                       & \textbf{3.33}  & \underline{0.34}  & -     & -                         \\
                     &                      & CW                               & \textbf{7.68}  & \underline{3.21}  & -     & -                         \\
                     &                      & MIM                              & \textbf{5.23}  & \underline{0.51}  & -     & -                         \\ \cmidrule(l){2-7} 
                     & \multirow{6}{*}{50}  & \multicolumn{1}{l}{None (Clean)} & \underline{39.89}          & \textbf{46.80} & -     & -                         \\
                     &                      & PGD100                           & \textbf{9.77}  & \underline{0.59}  & -     & -                         \\
                     &                      & Square                           & \underline{28.39}          & \textbf{32.40} & -     & -                         \\
                     &                      & AutoAttack                       & \textbf{7.03}  & \underline{0.47}  & -     & -                         \\
                     &                      & CW                               & \textbf{14.14} & \underline{6.31}  & -     & -                         \\
                     &                      & MIM                              & \textbf{9.84}  & \underline{0.64}  & -     & -                         \\ \bottomrule
\end{tabular}}%
\label{table:main}
\end{table}

\subsection{Ablation Study on Gradient Regularization}
Eq. \ref{eq:bound_main} showed that the adversarial loss is upper-bounded by the normal loss, the gradient magnitude, and the curvature term. GUARD regularizes the curvature term while disregarding the gradient magnitude, which could theoretically reduce the upper bound of the loss as well. Here, we investigate the effect of regularizing gradient instead of curvature and present the results in Table \ref{table:ablation}. The results indicate that GUARD outperforms the gradient regularization alternatives, regardless of the regularization parameter.

\begin{table}[ht]
\centering
\footnotesize
\caption{Accuracy on ImageNette of original \sre, GUARD, and gradient regularization on \sre~with regularization parameters ($\lambda_g=10^{-4}, 10^{-3}, 10^{-2}, 0.1, 1$, where $\lambda_g$ is omitted in table columns for brevity). AA stands for AutoAttack. The best results among all methods are highlighted in bold.
}
\setlength{\tabcolsep}{0.7mm}{
\begin{tabular}{@{}ccccccccc@{}}
\toprule
\multirow{2}{*}{IPC} & \multirow{2}{*}{Attack} & \multicolumn{7}{c}{Methods}                                           \\ \cmidrule(l){3-9} 
                     &                         & \sre & GUARD    & 
                     $10^{-4}$ & $10^{-3}$ & $10^{-2}$ & $0.1$ & $1$   \\ 
                     \midrule
\multirow{6}{*}{1}   & (Clean)            & 27.97 & \textbf{37.49} & 13.72   & 16.15   & 16.41   & 17.58  & 18.75 \\
                     & PGD100                  & 12.05 & \textbf{16.22} & 6.39    & 9.32    & 10.27   & 10.39  & 13.27 \\
                     & Square                  & 18.62 & \textbf{26.74} & 9.71    & 11.69   & 13.68   & 12.94  & 15.97 \\
                     & AA              & 12.12 & \textbf{15.81} & 6.32    & 9.35    & 10.17   & 10.42  & 13.25 \\
                     & CW                      & 20.38 & \textbf{29.14} & 7.62    & 11.06   & 11.64   & 11.31  & 14.47 \\
                     & MIM                     & 12.05 & \textbf{16.32} & 6.39    & 9.25    & 10.39   & 10.39  & 13.27 \\ \midrule
\multirow{6}{*}{10}  & (Clean)            & 42.42 & \textbf{57.93} & 44.82   & 41.81   & 42.80   & 43.34  & 40.31 \\
                     & PGD100                  & 4.76  & \textbf{23.87} & 15.39   & 14.47   & 15.41   & 13.68  & 16.92 \\
                     & Square                  & 22.77 & \textbf{44.07} & 34.06   & 31.80   & 34.73   & 31.85  & 31.03 \\
                     & AA              & 4.99  & \textbf{19.69} & 15.62   & 14.52   & 15.64   & 13.78  & 16.87 \\
                     & CW                      & 22.11 & \textbf{41.47} & 21.58   & 20.64   & 21.17   & 18.85  & 21.53 \\
                     & MIM                     & 4.76  & \textbf{21.80} & 15.41   & 14.60   & 15.34   & 13.66  & 17.41 \\ \midrule
\multirow{6}{*}{50}  & (Clean)            & 80.15 & \textbf{80.86} & 76.46   & 74.06   & 73.12   & 74.52  & -     \\
                     & PGD100                  & 12.30 & \textbf{41.42} & 35.54   & 35.36   & 33.48   & 29.17  & -     \\
                     & Square                  & 61.50 & \textbf{72.81} & 67.08   & 63.24   & 63.97   & 65.53  & -     \\
                     & AA              & 12.91 & \textbf{42.47} & 35.59   & 35.54   & 33.52   & 29.32  & -     \\
                     & CW                      & 53.42 & \textbf{58.67} & 46.06   & 45.43   & 44.05   & 41.02  & -     \\
                     & MIM                     & 12.43 & \textbf{43.23} & 35.61   & 35.39   & 33.66   & 29.07  & -     \\ \bottomrule
\end{tabular}}%
\label{table:ablation}
\end{table}

\section{Discussion}
\label{sec: discussion}
\subsection{Robustness Guarantee}
Due to the nature of dataset distillation, it is impossible to optimize the robustness of the final model with respect to the real dataset. Therefore, most approaches in this direction, including ours, have to optimize the adversarial loss of the model with respect to the distilled dataset. Unfortunately, there is always a distribution shift between the real and distilled datasets, which raises uncertainties about whether robustness on the distilled dataset will be effectively transferred when evaluated against the real dataset. Nevertheless, our theoretical framework offers assurances regarding this concern. A comparison between Eq. \ref{eq:exp_taylor_main} and Eq. \ref{eq:bound_main} reveals that the bounds of adversarial loss for real data and distilled data differ only by $L\sigma$. For a well-distilled dataset, $\sigma$ should be relatively small. We have thus demonstrated that the disparity between minimizing adversarial loss on the distilled dataset and on the real dataset is confined to this constant. This conclusion of our theory allows future robust dataset distillation methods to exclusively enhance robustness with respect to the distilled dataset, without worrying if the robustness can transfer well to the real dataset.

\subsection{Computational Overhead}
The structure of robust dataset distillation, as outlined in Eq. \ref{eq: formulation_robust}, inherently presents a tri-level optimization challenge. Typically, addressing such a problem could entail employing complex tri-level optimization algorithms, resulting in significant computational demands. One example of this is the integration of adversarial training within the distillation framework, which necessitates an additional optimization loop for generating adversarial examples within each iteration. However, GUARD's approach, as detailed in Eq. \ref{rloss_main}, introduces an efficient alternative. GUARD's regularization loss only requires an extra forward pass to compute the loss $\ell(\mathbf{x}+h\mathbf{z})$ within each iteration. Therefore, integrating GUARD's regularizer into an existing method does not significantly increase the overall computational complexity, ensuring that the computational overhead remains minimal. This efficiency is particularly notable given the computationally intensive nature of tri-level optimization in robust dataset distillation. 
In Table \ref{table:time}, we present a comparison of the time per iteration required for a tri-level optimization algorithm, such as the one used for embedded adversarial training, against the time required for GUARD. The findings show that GUARD is much more computationally efficient and has a lower memory usage as well.

\begin{table}[ht]
\centering
\fontsize{9pt}{11pt}\selectfont
\caption{Computation overhead of GUARD compared with embedded adversarial training. Experiments are performed on one NVIDIA A100 80GB PCIe GPU with batch size 32. We measure 5 times per iteration training time and report the average and standard deviation.}
\setlength{\tabcolsep}{4mm}{
\begin{tabular}{lllll}
\toprule
 Method & Time (s) / Iter & Peak Mem  \\
 \midrule
 GUARD & 0.007 $\pm$ 2.661$e^{-4}$ & 3851MiB \\
 Adv Training & 2.198 $\pm$ 6.735$e^{-4}$ & 4211MiB  \\
\bottomrule
\end{tabular}}
\label{table:time}
\end{table}

\subsection{Transferability}

Our investigation focuses on studying the effectiveness of the curvature regularizer within the \sre~framework. Theoretically, this method can be extended to a broad spectrum of dataset distillation methods. GUARD's application is feasible for any distillation approach that utilizes a model trained on the original dataset as a comparison target during the distillation phase — a strategy commonly seen across many dataset distillation techniques as noted in the Related Works section. This criterion is met by the majority of dataset distillation approaches, with the exception of those following the distribution matching approach, which may not consistently employ a comparison model~\cite{sachdeva2023survey}. This observation suggests GUARD's potential compatibility with a wide array of dataset distillation strategies. To demonstrate this, we explored two additional implementation of GUARD using DC~\cite{zhao2021dc} and CDA~\cite{yin2023cda} as baseline distillation methods. DC represents an earlier, simpler approach that leverages gradient matching for distillation purposes, whereas CDA is a more recent distillation technique, specifically designed for very large datasets. As shown in Table \ref{table:other}, GUARD consistently improves both clean accuracy and robustness across various dataset distillation methods.

\begin{table}[ht]
\centering
\fontsize{9pt}{11pt}\selectfont
\caption{Direct comparison of the original DC, \sre, and CDA methods with the addition of GUARD regularizer (marked by \textsuperscript{†}) on CIFAR10. The best results among each pair of compared methods are highlighted in bold.}
\setlength{\tabcolsep}{1mm}{
\begin{tabular}{@{}cccccccc@{}}
\toprule
\multirow{2}{*}{IPC} & \multirow{2}{*}{Attack} & \multicolumn{6}{c}{Methods}                                                   \\ \cmidrule(l){3-8} 
                     &                         & DC         & DC\textsuperscript{†}      & \sre & \sre\textsuperscript{†}   & CDA   & CDA\textsuperscript{†}   \\ \midrule
\multirow{6}{*}{1}   & None (Clean)            & 29.96       & \textbf{30.95} & 17.13 & \textbf{22.88} & 14.98 & \textbf{23.18} \\
                     & PGD100                  & 24.59      & \textbf{46.88} & 13.56 & \textbf{19.21} & 12.69 & \textbf{18.70}  \\
                     & Square                  & 24.72      & \textbf{48.56} & 13.75 & \textbf{19.55} & 12.84 & \textbf{19.17} \\
                     & AutoAttack              & \textbf{24.33}      & 14.99          & 13.43 & \textbf{18.91} & 12.63 & \textbf{18.42} \\
                     & CW                      & 24.58      & \textbf{15.19} & 13.52 & \textbf{18.95} & 12.62 & \textbf{18.52} \\
                     & MIM                     & \textbf{24.62}       & 15.27          & 13.57 & \textbf{19.22} & 12.69 & \textbf{18.71} \\ \midrule
\multirow{6}{*}{10}  & None (Clean)            & 45.38       & \textbf{46.83}  & 26.58 & \textbf{30.76} & 20.55 & \textbf{30.65} \\
                     & PGD100                  & 31.84      & \textbf{32.36} & 18.24 & \textbf{22.31} & 14.60 & \textbf{24.33} \\
                     & Square                  & \textbf{33.71}      & 33.54          & 19.99 & \textbf{24.16} & 15.93 & \textbf{25.66} \\
                     & AutoAttack              & 31.05      & \textbf{31.84} & 18.11 & \textbf{21.58} & 14.47 & \textbf{24.04} \\
                     & CW                      & 31.95      & \textbf{32.35} & 18.73 & \textbf{21.98} & 14.83 & \textbf{24.51} \\
                     & MIM                     & 31.89      & \textbf{32.37} & 18.25 & \textbf{22.35} & 14.62 & \textbf{24.34} \\ \midrule
\multirow{6}{*}{50}  & None (Clean)            & -          & -              & 43.96 & \textbf{44.05} & 36.32 & \textbf{43.05} \\
                     & PGD100                  & -          & -              & 24.74 & \textbf{33.12} & 21.58 & \textbf{33.02} \\
                     & Square                  & - & -              & 29.68 & \textbf{35.22} & 25.76 & \textbf{35.19} \\
                     & AutoAttack              & - & -              & 24.45 & \textbf{32.24} & 21.46 & \textbf{31.96} \\
                     & CW                      & - & -              & 26.09 & \textbf{32.67}  & 22.54 & \textbf{32.56} \\
                     & MIM                     & - & -              & 24.81 & \textbf{33.12} & 21.61 & \textbf{33.03} \\ \bottomrule
\end{tabular}}
\label{table:other}
\end{table}

\section{Conclusions}
\label{sec: conclusion}
Our work focuses on a novel perspective on dataset distillation by emphasizing its adversarial robustness characteristics. Upon reaching the theoretical conclusion that the adversarial loss of distilled datasets is bounded by the curvature, we proposed GUARD, a method that can be integrated into many dataset distillation methods to provide robustness against diverse types of attacks and potentially improve clean accuracy. Our theory also provided the insight that the optimization of robustness with respect to distilled and real datasets is differentiated only by a constant term, which may open up potentials for subsequent research in the field. Future work could explore the integration of robustness into more dataset distillation approaches as well as out-of-distribution settings. We hope our work contributes to the development of DD methods that are not only efficient but also robust, and will inspire further research in this area.

\bigskip
\bibliography{main}

\newpage
\newcommand{\sssec}[1]{\vspace*{0.05in}\noindent\textbf{#1}}
\setcounter{page}{1}
\setcounter{section}{0} 
\onecolumn
\maketitlesupplementary
\renewcommand{\thesection}{\Alph{section}}



\section{Proof of Proposition 1}
\label{proof}
\begin{figure*}[ht]
  \centering
    \includegraphics[width=14cm,height=14cm,keepaspectratio]{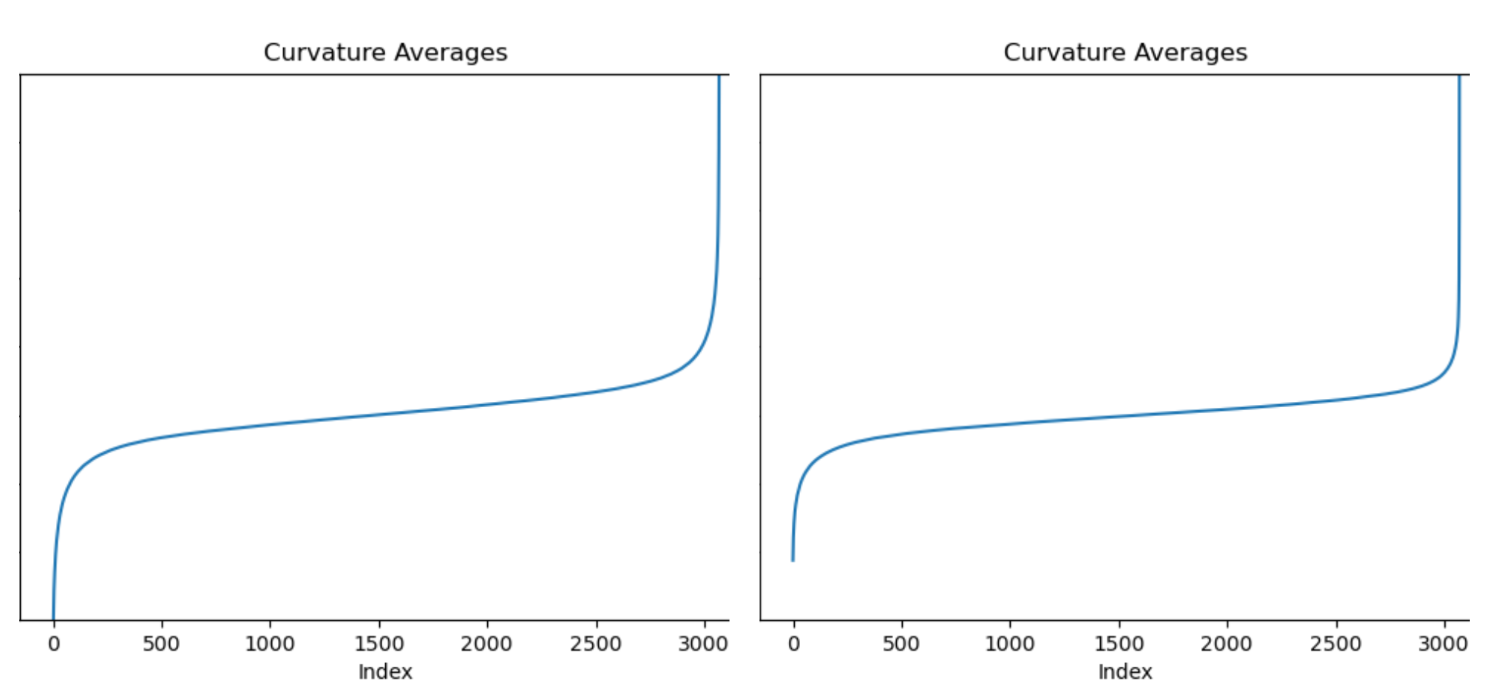}
  \caption{A comparison between the curvature profiles of a baseline dataset distillation method (left) and GUARD (right) in the form of sorted eigenvalues of the hessian}
  \label{figure3}
\end{figure*}

The adversarial loss of an arbitrary input sample $\mathbf{x}$ can be upper-bounded as below:
\begin{equation}
\begin{split}
\tilde{\ell}_\rho^{adv}(\mathbf{x}) & = \max_{\|\mathbf{v}\| \le \rho} \tilde{\ell}(\mathbf{x+v}) \\ &= \max_{\|\mathbf{v}\| \le \rho} \ell(\mathbf{x}) + {\nabla \ell(\mathbf{x})}^\top \mathbf{v} + \frac{1}{2} \mathbf{v^\top Hv} \\
&\le \max_{\|\mathbf{v}\| \le \rho} \ell(\mathbf{x}) + \|{\nabla \ell(\mathbf{x})}\| \|\mathbf{v}\| + \frac{1}{2} \lambda_1(\mathbf{x}) {\|\mathbf{v}\|}^2 \\
&= \ell(\mathbf{x}) + \|{\nabla \ell(\mathbf{x})}\| \rho + \frac{1}{2} \lambda_1(\mathbf{x}) \rho^2,
\label{app_eq1}
\end{split}
\end{equation}
where $\lambda$ is the largest eigenvalue of the Hessian $\mathbf{H(\ell(x))}$. \\
Taking expectation over the distribution of real data with class label $c$, denoted as $D_c$,
\begin{align}
\begin{split}
\label{eq:exp_taylor}
\mathop{\mathbb{E}}_{\mathbf{x} \sim D_c} \tilde{\ell}_\rho^{adv}(\mathbf{x}) \le 
\mathop{\mathbb{E}}_{\mathbf{x} \sim D_c}\ell(\mathbf{x}) + \rho \mathop{\mathbb{E}}_{\mathbf{x} \sim D_c} \|{\nabla \ell(\mathbf{x})}\| \\
+ \frac{1}{2} \rho^2 \mathop{\mathbb{E}}_{\mathbf{x} \sim D_c} \lambda_1(\mathbf{x}). 
\end{split}
\end{align}
With the assumption that $\tilde{\ell}(\mathbf{x})$ is convex, we know that $\tilde{\ell}_\rho^{adv}(\mathbf{x})$ is also convex, because $\forall \lambda \in [0, 1],$
\begin{equation}
\begin{split}
    &\tilde{\ell}_\rho^{adv}(\lambda \mathbf{x}_1 + (1-\lambda) \mathbf{x}_2) \\
    &= \max_{\|\mathbf{v}\| \le \rho} \tilde{\ell}(\lambda \mathbf{x}_1 + (1-\lambda) \mathbf{x}_2 + \mathbf{v}) \\
    &= \max_{\|\mathbf{v}\| \le \rho} \tilde{\ell}(\lambda (\mathbf{x}_1 + \mathbf{v})+ (1-\lambda) (\mathbf{x}_2 + \mathbf{v})) \\
    &\le \max_{\|\mathbf{v}\| \le \rho} \lambda \tilde{\ell}(\mathbf{x}_1 + \mathbf{v}) + (1-\lambda)\tilde{\ell}(\mathbf{x}_2 + \mathbf{v}) \\
    &\le \lambda \max_{\|\mathbf{v}\| \le \rho} \tilde{\ell}(\mathbf{x}_1 + \mathbf{v}) + (1-\lambda)\max_{\|\mathbf{v}\| \le \rho} \tilde{\ell}(\mathbf{x}_2 + \mathbf{v}) \\
    &= \lambda \tilde{\ell}_\rho^{adv}(\mathbf{x}_1) + (1-\lambda)\tilde{\ell}_\rho^{adv}(\mathbf{x}_2).
\end{split}
\end{equation}
Therefore, by Jensen’s Inequality,
\begin{align}
\label{eq:jensen}
   \tilde{\ell}_\rho^{adv}(\mathop{\mathbb{E}}_{\mathbf{x} \sim D_c} \mathbf{x}) \le \mathop{\mathbb{E}}_{\mathbf{x} \sim D_c} \tilde{\ell}_\rho^{adv}(\mathbf{x}).
\end{align}
Let $\mathbf{x}^\prime$ be a datum distilled from the training data with class label $c$. It should be close in distribution to that of the real data. Hence, we can assume the maximum mean discrepancy (MMD) between the distilled data and real data is bounded as $\|h(\mathbf{x}^\prime) - \mathbb{E}_{\mathbf{x} \sim D_c}h(\mathbf{x})\| \le \sigma$, where $h(\cdot)$ is a feature extractor. If $h(\cdot)$ is invertible, then $\mathcal{L}_\rho^{adv}(\cdot)=\tilde{\ell}_\rho^{adv}(h^{-1}(\cdot))$ is a function defined on the feature space. We assume that $\mathcal{L}_\rho^{adv}(\cdot)$ is $L$-Lipschitz, it follows that
\begin{align}
    \mathcal{L}_\rho^{adv}(h(\mathbf{x}^\prime)) \le \mathcal{L}_\rho^{adv}(\mathop{\mathbb{E}}_{\mathbf{x} \sim D_c} h(\mathbf{x})) + L\sigma.
\end{align}
If we add the assumption that $h(\cdot)$ is linear, $\mathop{\mathbb{E}}_{\mathbf{x} \sim D_c} h(\mathbf{x}) = h(\mathop{\mathbb{E}}_{\mathbf{x} \sim D_c} \mathbf{x})$, then
\begin{align}
\label{eq:mmd}
    \tilde{\ell}_\rho^{adv}(\mathbf{x}^\prime) \le \tilde{\ell}_\rho^{adv}(\mathop{\mathbb{E}}_{\mathbf{x} \sim D_c} \mathbf{x}) + L\sigma.
\end{align}
Combining ~Eq. \ref{eq:exp_taylor}, Eq. \ref{eq:jensen}, Eq. \ref{eq:mmd}, we get
\begin{align}
\begin{split}
\label{eq:bound}
    \tilde{\ell}_\rho^{adv}(\mathbf{x}^\prime) \le \mathop{\mathbb{E}}_{\mathbf{x} \sim D_c} \ell(\mathbf{x}) + \rho \mathop{\mathbb{E}}_{\mathbf{x} \sim D_c} \|\nabla \ell(\mathbf{x})\| \\
    + \frac{1}{2} \rho^2 \mathop{\mathbb{E}}_{\mathbf{x} \sim D_c} \lambda_1(\mathbf{x}) + L\sigma.
\end{split}
\end{align}
\paragraph{Discussion} The inequality in Eq. \ref{app_eq1} is an equality if and only if the direction of the gradient is the same as the direction of $\lambda_1$. Previous work has empirically shown that the two directions have a large cosine similarity in the input space of neural networks. Our assumption about the Lipschitz continuity of $\mathcal{L}_\rho^{adv}(\cdot)$ is reasonable, as recent work has shown improved estimation of the Lipschitz constant of neural networks in a wide range of settings~\cite{khromov2023fundamental}. Although our assumptions about the convexity of $\tilde{\ell}(\mathbf{x})$ and the linearity of $h(\cdot)$ is relatively strong, it still reflects important aspects of reality, as our experiment in Table \ref{table:main} has shown that reducing the curvature term in r.h.s of Eq. \ref{eq:bound} effectively improves the robustness of models trained on distilled data. 
Moreover, in Fig. \ref{figure3} we plot the distribution of eigenvalues of the real data samples on the loss landscape of a model trained on standard distilled data and a model trained on robust distilled data from our GUARD method, respectively. GUARD corresponds to a flatter curve of eigenvalue distribution, indicating that the loss landscape becomes more linear after our regularization.

\section{Additional Results}
In this section of the appendix, we provide supplementary results from our experiments. Table~\ref{table:app-main} presents a detailed comparison of the effects of various adversarial attacks on GUARD, \sre, MTT, and TESLA, which were excluded from the main paper due to space constraints. The results further highlight GUARD's improved adversarial robustness, mai=ntaining a positive trend of being much better than other methods.

Furthermore, Table~\ref{table:app-compute} showcases additional comparisons illustrating the computational efficiency of GUARD. The results demonstrate that GUARD consistently achieves significantly faster runtimes per iteration compared to adversarial training.

\begin{table}[ht]
\centering
\fontsize{9pt}{11pt}\selectfont
\caption{Evaluation of different dataset distillation methods under adversarial attacks on ImageNette and TinyImageNet, under the 1 IPC setting. The best results among all methods are highlighted in bold, second best are underlined.}
\setlength{\tabcolsep}{1mm}{
\begin{tabular}{@{}ccccccc@{}}
\toprule
\multirow{2}{*}{Dataset}      & \multirow{2}{*}{IPC} & \multirow{2}{*}{Attack} & \multicolumn{4}{c}{Methods}            \\ \cmidrule(l){4-7} 
 &  &            & GUARD          & SRe2L & MTT            & \multicolumn{1}{l}{TESLA} \\ \midrule
\multirow{5}{*}{Imagenette}   & \multirow{5}{*}{1}   & PGD100                  & 16.22 & 12.05 & \textbf{18.60} & 24.68 \\
 &  & Square     & \textbf{26.74} & 18.62 & 1.40           & 22.21                     \\
 &  & AutoAttack & \textbf{15.81} & 12.12 & 1.40           & 22.16                     \\
 &  & CW         & \textbf{29.14} & 20.38 & 1.40           & 22.26                     \\
 &  & MIM        & 16.32          & 12.05 & \textbf{18.60} & 24.68                     \\ \midrule
\multirow{5}{*}{TinyImagenet} & \multirow{5}{*}{1}   & PGD100                  & 1.48  & 0.83  & \textbf{3.26}  & 3.64  \\
 &  & Square     & \textbf{3.26}  & 2.44  & 1.76           & 2.83                      \\
 &  & AutoAttack & \textbf{0.74}  & 0.66  & 1.76           & 2.78                      \\
 &  & CW         & \textbf{1.54}  & 1.18  & 1.76           & 2.82                      \\
 &  & MIM        & 1.49           & 0.88  & \textbf{3.28}  & 3.64                      \\ \bottomrule
\end{tabular}}
\label{table:app-main}
\end{table}

\begin{table}[ht]
\centering
\fontsize{9pt}{11pt}\selectfont
\caption{Relative slowdown introduced by adding GUARD versus adversarial training to \sre~in terms of average runtime per iteration, tested on three variations of the ImageNette dataset with image sizes of 160x160px, 320x320px, and the original size, using two different graphics cards.}
\setlength{\tabcolsep}{0.5mm}{
\begin{tabular}{@{}ccccccc@{}}
\toprule
\multirow{2}{*}{Method} & \multicolumn{3}{c}{A100}              & \multicolumn{3}{c}{RTX4090}           \\
                        & 160px & 320px & Original & 160px & 320px & Original \\ \midrule
GUARD                   & 4.63x     & 3.96x     & 4.54x         & 2.37x     & 3.26x     & 2.81x         \\
Adv. Training           & 49.83x    & 49.96x    & 53.40x        & 35.14x    & 53.22x    & 41.23x        \\ \bottomrule
\end{tabular}}
\label{table:app-compute}
\end{table}

\section{Effect of GUARD on Images}
\begin{figure*}[ht!]
\centering

\newcommand{\imagewidth}{0.18\textwidth} 

\begin{subfigure}[t]{\imagewidth}
    \includegraphics[width=\linewidth]{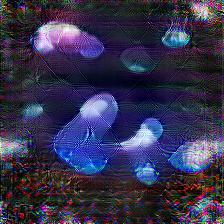}
    \caption*{}
\end{subfigure}
\hfill
\begin{subfigure}[t]{\imagewidth}
    \includegraphics[width=\linewidth]{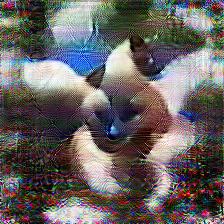}
    \caption*{}
\end{subfigure}
\hfill
\begin{subfigure}[t]{\imagewidth}
    \includegraphics[width=\linewidth]{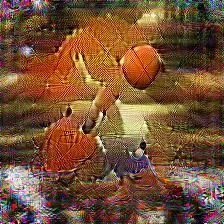}
    \caption*{}
\end{subfigure}
\hfill
\begin{subfigure}[t]{\imagewidth}
    \includegraphics[width=\linewidth]{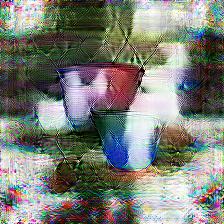}
    \caption*{}
\end{subfigure}
\hfill
\begin{subfigure}[t]{\imagewidth}
    \includegraphics[width=\linewidth]{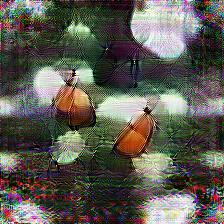}
    \caption*{}
\end{subfigure}
\vspace{-8mm}
\caption*{GUARD}
\vspace{3mm}

\begin{subfigure}[t]{\imagewidth}
    \includegraphics[width=\linewidth]{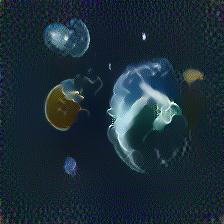}
    \caption*{Jellyfish}
\end{subfigure}
\hfill
\begin{subfigure}[t]{\imagewidth}
    \includegraphics[width=\linewidth]{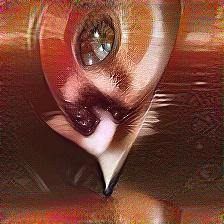}
    \caption*{Siamese cat}
\end{subfigure}
\hfill
\begin{subfigure}[t]{\imagewidth}
    \includegraphics[width=\linewidth]{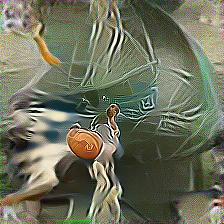}
    \caption*{Basketball}
\end{subfigure}
\hfill
\begin{subfigure}[t]{\imagewidth}
    \includegraphics[width=\linewidth]{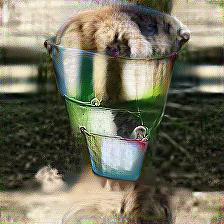}
    \caption*{Bucket}
\end{subfigure}
\hfill
\begin{subfigure}[t]{\imagewidth}
    \includegraphics[width=\linewidth]{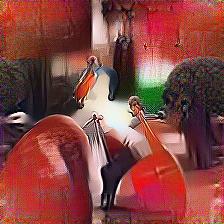}
    \caption*{Cello}
\end{subfigure}
\vspace{-3mm}
\caption*{SRe$^2$L}
\vspace{5mm}
 
\begin{subfigure}[t]{\imagewidth}
    \includegraphics[width=\linewidth]{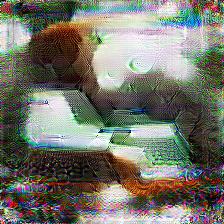}
    \caption*{}
\end{subfigure}
\hfill
\begin{subfigure}[t]{\imagewidth}
    \includegraphics[width=\linewidth]{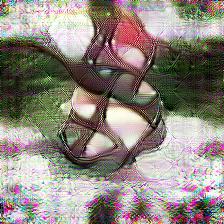}
    \caption*{}
\end{subfigure}
\hfill
\begin{subfigure}[t]{\imagewidth}
    \includegraphics[width=\linewidth]{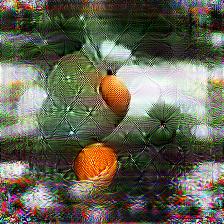}
    \caption*{}
\end{subfigure}
\hfill
\begin{subfigure}[t]{\imagewidth}
    \includegraphics[width=\linewidth]{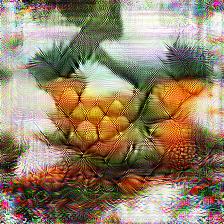}
    \caption*{}
\end{subfigure}
\hfill
\begin{subfigure}[t]{\imagewidth}
    \includegraphics[width=\linewidth]{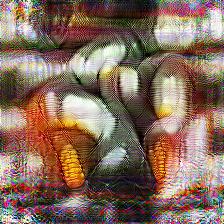}
    \caption*{}
\end{subfigure}
\vspace{-8mm}
\caption*{GUARD}
\vspace{3mm}

\begin{subfigure}[t]{\imagewidth}
    \includegraphics[width=\linewidth]{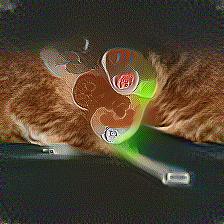}
    \caption*{Laptop}
\end{subfigure}
\hfill
\begin{subfigure}[t]{\imagewidth}
    \includegraphics[width=\linewidth]{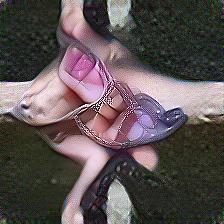}
    \caption*{Sandal}
\end{subfigure}
\hfill
\begin{subfigure}[t]{\imagewidth}
    \includegraphics[width=\linewidth]{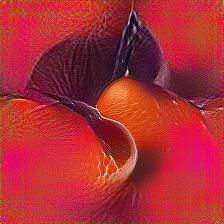}
    \caption*{Orange}
\end{subfigure}
\hfill
\begin{subfigure}[t]{\imagewidth}
    \includegraphics[width=\linewidth]{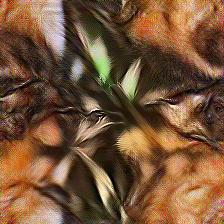}
    \caption*{Pineapple}
\end{subfigure}
\hfill
\begin{subfigure}[t]{\imagewidth}
    \includegraphics[width=\linewidth]{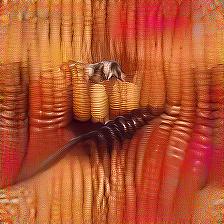}
    \caption*{Corn}
\end{subfigure}
\vspace{-3mm}
\caption*{SRe$^2$L}

\caption{Comparative visualization of distilled images from GUARD and \sre~with 1 ipc setting on ImageNet.}
\label{viz_guard_sre}
\end{figure*}

In this section, we provide a detailed comparison between synthetic images produced by GUARD and those generated by other methods. In Fig. \ref{viz_guard_sre}, we showcase distilled images from GUARD (utilizing \sre~as a baseline) alongside images from \sre.

Our comparative analysis reveals that the images generated by the GUARD method appear to have more distinct object outlines when compared with those from the baseline \sre~method. This improved definition of objects may facilitate better generalization in subsequent model training, which could offer an explanation for the observed increases in clean accuracy.

Additionally, our synthetic images exhibit a level of high-frequency noise, which bears similarity to the disruptions introduced by adversarial attacks. While visually subtle, this attribute might play a role in enhancing the resilience of models against adversarial inputs, as training on these images could prepare the models to handle unexpected perturbations more effectively. This suggests that the GUARD method could represent a significant advancement in the creation of synthetic datasets that promote not only visual fidelity but also improved robustness in practical machine learning applications.

\section{Algorithm of GUARD with Optimization-based Distillation Methods}
\label{sec: alg}
In our main paper, we explained how GUARD can be easily intergrated into \sre~by incorporating the regularizer into the model's training loss during the squeeze (pre-training) step. We later demonstrated that GUARD could also be integrated into other distillation methods, such as DC \cite{zhao2021dc}. However, unlike \sre, DC lacks a pre-training phase; instead, the model's training and distillation occur simultaneously, making the integration less straightforward.

Therefore, we present the GUARD algorithm using DC as the baseline method in Alg. \ref{alg:1}. For each outer iteration $k$, we sample a new initial weight from some random distribution of weights to ensure the synthetic dataset can generalize well to a range of weight initializations. After, we iteratively sample a minibatch pair from the real dataset and the synthetic dataset and compute the loss over them on a neural network with the weights $\boldsymbol{\theta}_t$. We compute the regularized loss on real data through Eq. \ref{rloss_main}. Finally, we compute the gradient of the losses w.r.t. $\boldsymbol{\theta}$, and update the synthetic dataset through stochastic gradient descent on the distance between the gradients. At the end of each inner iteration $t$, we update the weights $\boldsymbol{\theta}_{t+1}$ using the updated synthetic dataset.

\RestyleAlgo{ruled}
\begin{algorithm*}[ht]
\caption{Algorithm of GUARD (based on DC)}
\label{alg:1}
\KwData{{$\T$: Training set}; {$\mS$: initial synthetic dataset with $C$ classes}; $p(\theta_0)$: initial weights distribution; {$\phi_\theta$: neural network}; $K$: number of outer-loop steps; {$T$: number of inner-loop steps}; $\varsigma_\theta$: number of steps for updating weights; $\varsigma_S$: number of steps for updating synthetic samples; { $\eta_\theta$: learning rate for updating weights}; {$\eta_S$: learning rate for updating synthetic samples}; {$D$: gradient distance function}; {$h$: discretization step}; $\lambda$: strength of regularization}

\ForEach{{\upshape outer training step $k = 1$ to $K$}}{
    Sample initial weight $\theta_0 \sim p(\theta_0)$ \;
    \ForEach{{\upshape inner training step $t = 1$ to $T$}}{
        \ForEach{{\upshape class $c = 1$ to $C$}}{
            Sample $\omega\sim\Omega$ and a minibatch pair $B^{\T}_c\sim\T$ and $B^{\mS}_c\sim\mS$\; 
            Compute loss on synthetic data  $\ELL^\mS_c = \frac{1}{|B^{\mS}_c|} \sum_{(\bs,\by)\in B^{\mS}_c}\ell(\phi_{\theta_t}(\bs), \by)$ \;
            Compute loss on real data $\ELL^\T_c = \frac{1}{|B^{\T}_c|} \sum_{(\bx,\by)\in B^{\T}_c}\ell(\phi_{\theta_t}(\bx), \by)$\;
            Compute $z=\frac{\nabla\ell(\phi_{\theta_t}(\bs), \by)}{\|\nabla\ell(\phi_{\theta_t}(\bs, \by))\|}$\;
            Compute loss on perturbed real data $\ELL^{\T_z}_c = \frac{1}{|B^{\T}_c|} \sum_{(\bx,\by)\in B^{\T}_c}\ell(\phi_{\theta_t}(\bx + hz), \by)$\;
            Compute regularizer $\mathcal{R} = \nabla_\theta\ELL^{\T_z}_c(\theta_t)-\nabla_\theta\ELL^\T_c(\theta_t)$\;
            Compute regularized loss on real data $\ELL^{\T_\mathcal{R}}_c = \ELL^\T_c + \lambda\mathcal{R}$\;
            Update $\mS_c\leftarrow\texttt{sgd}_\mS (D(\nabla_\theta\ELL^\mS_c (\theta_t), \nabla_\theta\ELL^{\T_\mathcal{R}}_c (\theta_t)), \varsigma_\mS, \eta_\mS)$\;
        }
        Update $\theta_{t+1}\leftarrow\texttt{sgd}_\theta (\ELL(\theta_t, \mS), \varsigma_\theta, \eta_\theta)$ \;
    }
}
\KwResult{robust condensed dataset $\mS$}
\end{algorithm*}
\end{document}